# Registration of Brain Images using Fast Walsh Hadamard Transform

D.Sasikala [1] and  R.Neelaveni [2]
[1] Research Scholar, Assistant Professor,
Bannari Amman Institute of Technology, Sathyamangalam.
Tamil Nadu-638401.
Email address : anjansasikala@gmail.com

[2] Assistant Professor,
PSG College of Technology,   Coimbatore,
Tamil Nadu -641004.
Email address : rn64asok@yahoo.co.in

## Abstract

A lot of image registration techniques have been developed with great significance for data analysis in medicine, astrophotography, satellite imaging and few other areas. This work proposes a method for medical image registration using Fast Walsh Hadamard transform. This algorithm registers images of the same or different modalities. Each image bit is lengthened in terms of Fast Walsh Hadamard basis functions. Each basis function is a notion of determining various aspects of local structure, e.g., horizontal edge, corner, etc. These coefficients are normalized and used as numerals in a chosen number system which allows one to form a unique number for each type of local structure. The experimental results show that Fast Walsh Hadamard transform accomplished better results than the conventional Walsh transform in the time domain. Also Fast Walsh Hadamard transform is more reliable in medical image registration consuming less time.

## I. INTRODUCTION

Digital image processing is developing the ultimate machine that could perform the visual functions of all. It is a rapidly evolving field with growing applications in many areas of science and engineering. The main criterion of registration is to fuse the sets of data with the variations if any or with their similarities into a single data. These sets of data are acquired by sampling the same scene or object at different times or from different perspectives, in different co-ordinate systems. The purpose of registration is to visualize a single data merged with all the details about these sets of data obtained at different times or perspectives or co-ordinate systems. Such data is very essential in medicine for doctors to plan for surgery. The most common and important classes of image analysis algorithm with medical applications [1,3] are image registration and image segmentation. In Image analysis technique, the same input gives out somewhat detail description of the scene whose image is being considered. Hence the image analysis algorithms perform registration as a part of it towards producing the description. In single subject analysis, the statistical analysis is done either before or after registration. But in group analyses, it is done after registration.

Generally registration is the most difficult task, as aligning images to overlap the common features and differences if any are to be emphasized for immediate visibility to the naked eye. There is no general registration [1-17] algorithm, which can work reasonably well for all images. A suitable registration algorithm for the particular problem must be chosen or developed, as they are adhoc in nature. The algorithms can be incorporated explicitly or implicitly or even in the form of various parameters. This step determines the success or failure of image analysis. This technique may be classified based on four different aspects given as follows: (i) the feature selection (extracting features from an image) using their similarity measures and a correspondence basis, (ii) the transformation function, (iii) the optimization procedure, and (iv) the model for processing by interpolation.

Amongst the numerous algorithms developed for image registration so far, methods based on image intensity values are particularly excellent as they are simple to automate as solutions to optimization problems. Pure translations, for example, can be calculated competently, and universally, as the maxima of the cross correlation function between two images [11] [15] [17]. Additional commands such as rotations, combined with scaling, shears, give rise to nonlinear functions which must be resolved using iterative nonlinear optimization methods [11].

In the medical imaging field, image registration is regularly used to combine the complementary and synergistic information of images attained from different modalities. A problem when registering image data is that one does not have direct access to the density functions of the image intensities. They must be estimated from the image data. A variety of image registration techniques have been used for successfully registering images that are unoccluded and generally practiced with the use of Parzen windows or normalized frequency histograms [12].

The work proposed in this paper uses Fast Walsh Hadamard Transform (FWHT) [18, 19] for image registration. The coefficients obtained are normalized to determine a unique number which in turn represents the digits in a particular range. The experiments conducted on clinical images show that proposed algorithm performed well than the conventional Walsh Transform (WT) method in medical image registration. In addition, this paper provides a comparative analysis of FWHT and WT in Medical image registration.

The remainder of the paper is ordered as follows. Section 2 provides an overview on the related work for image registration. Section 3 explains WT in image registration. Section 4 describes the proposed approach for image registration using FWHT. Section 5 illustrates the experimental results to prove the efficiency of the proposed approach in image registration and Section 6 concludes the paper with a discussion.

## II. Related Work

Many discussions have been carried out previously on Image Registration. This section of paper provides a quick look on the relevant research work in image registration.

An automatic scheme using global optimization technique for retinal image registration was put forth by Matsopoulos et al. in [1]. A robust approach that estimates the affine transformation parameters necessary to register any two digital images misaligned due to rotation, scale, shear, and translation was proposed by Wolberg and Zokai in [2]. Zhu described an approach by cross-entropy optimization in [3]. Jan Kybic and Michael Unser together put forth an approach for fast elastic multidimensional intensity-based image registration with a parametric model of the deformation in [4]. Bentoutou et al. in [5] offered an automatic image registration for applications in remote sensing. A novel approach that addresses the range image registration problem for views having low overlap and which may include substantial noise for image registration was described by Silva et al. in [6]. Matungka et al. proposed an approach that involved Adaptive Polar Transform (APT) for Image registration in [7, 10]. A feature-based, fully non supervised methodology dedicated to the fast registration of medical images was described by Khaissidi et al. in [8]. Wei Pan et al. in [9] proposed a technique for image registration using Fractional Fourier Transform (FFT).

## I. Walsh Transform

Orthogonal transforms expand an image into sets of orthogonal basis images each of which represents a type of local structure. Examples are the Walsh, Haar [13], etc. The coefficients of such an extension point toward the effectiveness of the occurrence of the similar structure at the particular position. If these coefficients are normalized by the dc coefficient of the expansion, i.e., the local average gray value of the image, then they measure purely the local structure independent of modality. Walsh basis functions correspond to local structure, in the form of positive or negative going horizontal or vertical edge, corner of a certain type, etc. Registration schemes based on wavelet coefficient matching do not present a general mechanism of combining the matching results across different scales.

Two images $I_1$ and $I_2$, $I_1$ is assumed as reference image whereas $I_2$ represent an image that has to be deformed to match $I_1$. First, consider around each pixel, excluding border pixels, a 3X3 neighborhood and compute from it, the nine Walsh coefficients (3X3 WT of a 3X3 image patch). If 'f' is the input image, the matrix of coefficients 'g' computed for it using equation (1),

$$g = (W^{-1})^T . fW^{-1} \qquad (1)$$

Matrix contains the coefficients of the expansion of the image, in terms of the basis images as in Figure.1 (a) formed by taking the vector outer products of the rows of matrix W[13]. Coefficients are denoted by $a_{00}$, $a_{01}$, $a_{02}$, $a_{10}$, $a_{11}$, $a_{12}$, $a_{20}$, $a_{21}$, $a_{22}$, in a matrix form as in Figure. 1(b) and take the value in the range [0, 9]. $\alpha_{ij}$ is normalization given in equation (2) makes the method robust to global levels of change of illumination. $a_{00}$ coefficient is the local average gray value of the image, $a_{ij}$ constructs coefficients that describes the local structure.

$$\alpha_{ij} = a_{ij} / a_{00} \qquad (2)$$

However, the information having dense features and rigid body transformation allows for plenty of redundancy in the system and makes it robust to noise and bad matches of individual pixels which effectively represent lack of local information. Construct a unique number out of eight numbers using these numbers as the digits of the unique number. The number of levels depends on the number system adopted. For decimal system, the normalized coefficients are quantized that taking integer values in the range [0, 9].

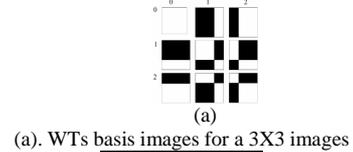

(a). WTs basis images for a 3X3 images

| $a_{00}$ | $a_{01}$ | $a_{02}$ |
|----------|----------|----------|
| $a_{10}$ | $a_{11}$ | $a_{12}$ |
| $a_{20}$ | $a_{21}$ | $a_{22}$ |

(b). Nine coefficients in matrix form
Figure 1. Walsh Transformation

In Figure 1(a) the coefficients along the first row and the first column are of equal importance, as they measure the presence of a vertical or a horizontal edge, respectively. The remaining four coefficients measure the presence of a corner. The following ordering of coefficients are used in images,

Ordering IA $\alpha_{01}$, $\alpha_{10}$, $\alpha_{20}$, $\alpha_{02}$, $\alpha_{11}$, $\alpha_{21}$, $\alpha_{12}$, $\alpha_{22}$

Ordering IB $\alpha_{10}$, $\alpha_{01}$, $\alpha_{02}$, $\alpha_{20}$, $\alpha_{11}$, $\alpha_{12}$, $\alpha_{21}$, $\alpha_{22}$

Ordering IIA $\alpha_{22}$, $\alpha_{21}$, $\alpha_{12}$, $\alpha_{11}$, $\alpha_{02}$, $\alpha_{20}$, $\alpha_{10}$, $\alpha_{01}$

Ordering IIB $\alpha_{22}$, $\alpha_{12}$, $\alpha_{21}$, $\alpha_{11}$, $\alpha_{20}$, $\alpha_{02}$, $\alpha_{01}$, $\alpha_{10}$

## II. Proposed Approach

### A. Fast Walsh Hadamard Transform

A fast transform algorithm is seen as a sparse factorization of the transform matrix, and refers to each factor as a stage. The proposed algorithms have a regular interconnection pattern between stages, and consequently, the inputs and outputs for each stage are addressed from or to the same positions, and the factors of the decomposition, the stages, have the property of being equal between them. The 2X2 Hadamard matrix is defined as $H_2$ is given in equation (3)

$$H_2 = \begin{bmatrix} 1 & 1 \\ 1 & -1 \end{bmatrix} \qquad (3)$$

A set of radix-R factorizations in terms of identical sparse matrices rapidly obtained from the WHT property that relates the matrix H with its inverse and is given in equation (4),

$$H_{R^n} = R^n (H_{R^n})^{-1} \qquad (4)$$

Where $H_{R^n}$ = radix-R Walsh Hadamard transform;

$R^n$ = radix-R factorizations;

n = input element;

The FWHT is utilized to obtain the local structure of the images. This basis function can be effectively used to obtain the digital numbers in the sense of coefficients [18] [19]. If these coefficients are normalized by the dc coefficient of the expansion, i.e., the local average gray value of the image, then they measure purely the local structure independent of modality. These numbers are then normalized to obtain the unique number that is used as feature for image registration. The implementation of FWHT readily reduces the time consumption for medical image registration when comparing the same with conventional WT technique for image registration.

## III. EXPERIMENTAL RESULT

A series of experiments is performed using medical images. The tests are performed using different images of different sizes. A set of CT and magnetic resonance (MR) medical images which depict the head of the same patient is considered. The original size of these images is given as pixels. In order to remove the background parts and the head outline, the original images are cropped, creating sub-images of different dimension pixels.

In probability theory and information theory, (sometimes known as transinformation) Mutual Information between two discrete random variables is defined as the amount of information shared between the two random variables. It is a dimensionless quantity with units of bits and can be the reduction in uncertainty. High MI indicates a large reduction in uncertainty; low MI indicates a small reduction; and zero MI between two random variables means the variables are independent.

Mutual Information

$$I(X;Y) = \sum_{y \in Y} \sum_{x \in X} p(x,y) \log\left(\frac{p(x,y)}{p_1(x)\,p_2(y)}\right), \quad (5)$$

- $X$ and $Y$ - Two discrete random variables.
- $p(x,y)$ - Joint probability distribution function of $X$ and $Y$.
- $p1(x)$ and $p2(y)$ - Marginal probability distribution functions of $X$ and $Y$ respectively.

The Correlation Coefficient is from statistics, is a measure of how well the predicted values from a forecast model "fit" with the real-life-data. If there is no relationship between the predicted values and actual values the CC is very low. As the strength of the relationship between the predicted values and actual values increases, so does the CC. Thus higher the CC the better it is.

Correlation Coefficient

$$C(t,s;\theta) = \frac{\sum_x \sum_y [I_1^{new}(x,y) - \overline{I_1^{new}(x,y)}][I_2^{new}(x\cos\theta - y\sin\theta - t, x\sin\theta + y\cos\theta - s) - \overline{I_2^{new}(x,y)}]}{\sqrt{\sum_x \sum_y [I_1^{new}(x,y) - \overline{I_1^{new}(x,y)}]^2}\sqrt{\sum_x \sum_y [I_2^{new}(x\cos\theta - y\sin\theta - t, x\sin\theta + y\cos\theta - s) - \overline{I_2^{new}(x,y)}]^2}} \quad (6)$$

$I_1^{new}, I_2^{new}$ - Two new images that differ from each other by rotation and translation only.

$t, s$ - Shifting parameters between the two images.

$\theta$ - Rotation angle.

$\overline{I_1^{new}(x,y)}, \overline{I_2^{new}(x,y)}$ - Average structure value of the pixels in the overlapping parts of images $I_1^{new}(x,y), I_2^{new}(x,y)$ respectively.

**(i) CT Sagittal Image – 432 x 427 – 41k JPEG , 36.3kB**

During image registration, Figure 2,(a) the registered image of base 1 is same for both WT & FWHT. Also Figure 2.(b) shows that base 1 of both WT & FWHT gives the same difference in images.

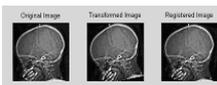

a) Registered Image obtained for Base 1 using WT & FWHT

b) Difference in images obtained for Base 1 using WT & FWHT
Figure 2. Images obtained for Base 1 using WT & FWHT

Figure 3.(a) is the registered image of base 2 for WT. Figure 3.(b) gives the difference in images. Figure 4,(a) is the registered image of base 2 for FWHT. Figure 4.(b) shows that base 2 of FWHT gives the difference in images. Both the results are different from each other. By analyses it proves that FWHT is better when compared to the results of WT.

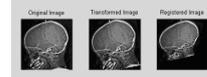

a) Registered Image obtained for Base 2 using WT

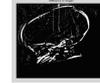

b) Difference in images obtained for Base 2 using WT
Figure 3. Images obtained for Base 2 using WT

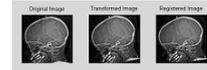

a) Registered Image obtained for Base 2 using FWHT

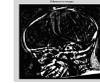

b) Difference in images obtained for Base 2 using FWHT
Figure 4. Images obtained for Base 2 using FWHT

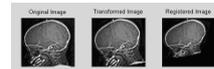

a) Registered Image obtained for Base 5 using WT

b) Difference in images obtained for Base 5 using WT
Figure 5. Images obtained for Base 5 using WT

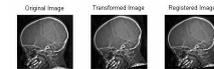

a) Registered Image obtained for Base 5 using FWHT

b) Difference in images obtained for Base 5 using FWHT
Figure 6. Images obtained for Base 5 using FWHT

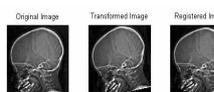

a) Registered Image obtained for Base 10 using WT & FWHT

b) Difference in images obtained for Base10 using WT& FWHT
Figure 7. Images obtained for Base 10 using WT & FWHT

**.(ii)    MRI T1-Registered**
 –      **Sagittal Image 400 x 400 –24k JPEG, 42.1kB and Frontal Image 400 x 400 – 11k JPEG, 30.9kB**

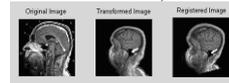

a) Registered Image obtained for Base 1 using WT & FWHT

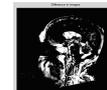

b) Difference in images obtained for Base1 using WT& FWHT
Figure 8. Images obtained for Base 1 using WT& FWHT

**MRI T1-Registered**
– **Frontal Image 400 x 400 – 11k JPEG, 30.9KB and
Sagittal Image 400 x 400 –24k JPEG, 42.1kB.**

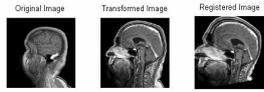

a) Registered Image obtained for Base 1 using WT

b)Difference in images obtained for Base1 using WT
Figure 9. Images obtained for Base 1 using WT

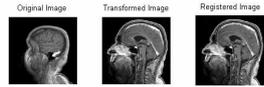

a) Registered Image obtained for Base 1 using FWHT

b)Difference in images obtained for Base1 using FWHT
Figure 10. Images obtained for Base 1 using FWHT

**MRI T1-Registered**
– **Sagittal Image 400 x 400 –24k JPEG, 42.1kB and
Frontal Image 400 x 400 – 11k JPEG, 30.9kB**

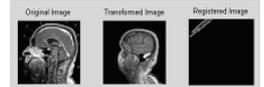

a) Registered Image obtained for Base 2 using WT

b)Difference in images obtained for Base2 using WT
Figure 11. Images obtained for Base 2 using WT

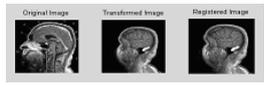

a) Registered Image obtained for Base2 using FWHT

b)Difference in images obtained for Base2 using FWHT
Figure 12. Images obtained for Base 2 using FWHT

**MRI T1-Registered**
– **Frontal Image 400 x 400 – 11k JPEG, 30.9kB and
Sagittal Image 400 x 400 –24k JPEG, 42.1kB.**

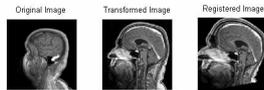

a) Registered Image obtained for Base 2 using WT

b)Difference in images obtained for Base2 using WT
Figure 13. Images obtained for Base 2 using WT

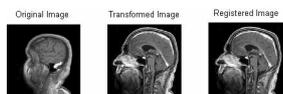

a) Registered Image obtained for Base2 using FWHT

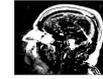

b)Difference in images obtained for Base2 using FWHT
Figure 14. Images obtained for Base 2 using FWHT

**MRI T1-Registered**
– **Sagittal Image 400 x 400 –24k JPEG, 42.1kB and
Frontal Image 400 x 400 – 11k JPEG, 30.9kB**

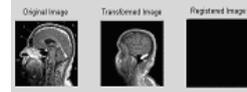

a) Registered Image obtained for Base5 using WT

b)Difference in images obtained for Base5 using WT
Figure 15. Images obtained for Base 5 using WT

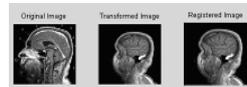

a) Registered Image obtained for Base5 using FWHT

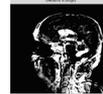

b)Difference in images obtained for Base5 using FWHT
Figure 16. Images obtained for Base 5 using FWHT

**MRI T1-Registered**
– **Frontal Image 400 x 400 – 11k JPEG, 30.9kB and
Sagittal Image 400 x 400 –24k JPEG, 42.1kB.**

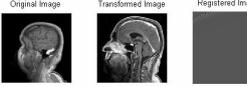

a) Registered Image obtained for Base5 using WT

b)Difference in images obtained for Base5 using WT
Figure 17. Images obtained for Base 5 using WT

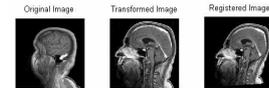

a) Registered Image obtained for Base5 using FWHT

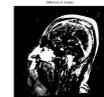

b)Difference in images obtained for Base5 using FWHT
Figure 18. Images obtained for Base 5 using FWHT

**MRI T1-Registered**
– **Sagittal Image 400 x 400 –24k JPEG, 42.1kB and
Frontal Image 400 x 400 – 11k JPEG, 30.9kB**

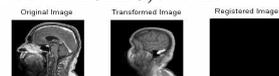

a) Registered Image obtained for Base10 using WT

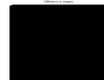

b)Difference in images obtained for Base10 using WT
Figure 19. Images obtained for Base 10 using WT

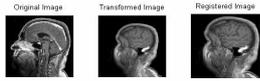

a) Registered Image obtained for Base10 using FWHT

**6.**

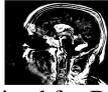

b)Difference in images obtained for Base10 using FWHT

Figure 20. Images obtained for Base10 using FWHT

**7.**

**MRI T1-Registered**
**– Frontal Image 400 x 400 – 11k JPEG, 30.9kB and**
**Sagittal Image 400 x 400 –24k JPEG, 42.1kB.**

**8.**

For Base 10 WT registration error occurred.

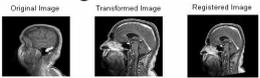

a) Registered Image obtained for Base10 using FWHT

**9.**

b)Difference in images obtained for Base10 using FWHT

Figure 21. Images obtained for Base10 using FWHT

**10.**

(iii) For the evaluation of the algorithm, 21 such sets of CT-MR image pairs are used.

(a) For base 1:

**11.**

For MRI T2-Registered –Sagittal Image 400 x 419 - 88.8kB the results of WT and FWHT are obtained that are almost similar. Figure 22.shows the pictorial outputs from the FWHT. Even the WT produces the same output as in Figure 22.

**12.**

Table 1 show the summary of all the results when a single ordering is taken into account using WT and FWHT in terms of MI. MI represents Mutual Information [16]. CC represents Correlation Coefficient. Figure 23.shows the performance comparison of WT and FWHT with respect to MI. Table 2 represents the summary of all results using conventional WT and FWHT in terms of CC. Table 3 indicates the time consumption for registering image using conventional WT and FWHT. Figure 24 represents the comparison of conventional WT and FWHT in terms of CC. Figure 25 represents the time consumption for registering image using conventional WT and FWHT.

**13.**

**1.**

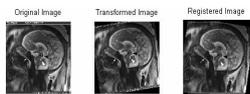

**2.**

**14.**

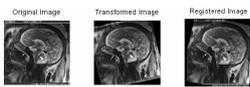

**3.**

**15.**

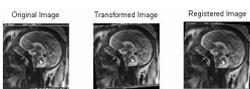

**4.**

**16.**

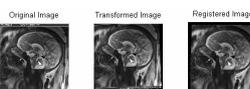

**5.**

**17.**

**18.**

**19.**

**20.**

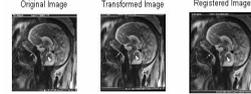
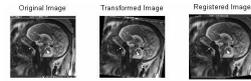
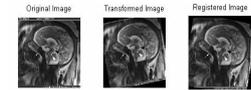
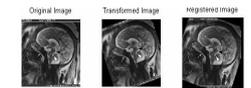
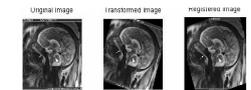
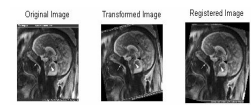
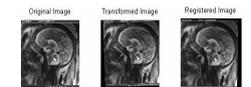
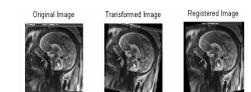
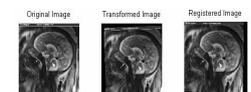
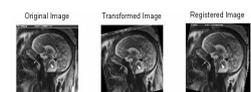
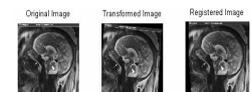
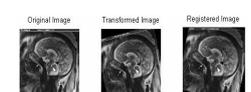
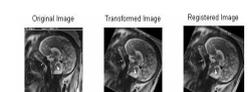
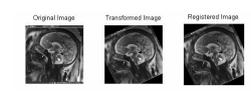
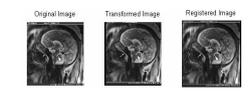

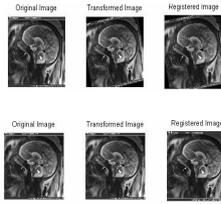

**21.**

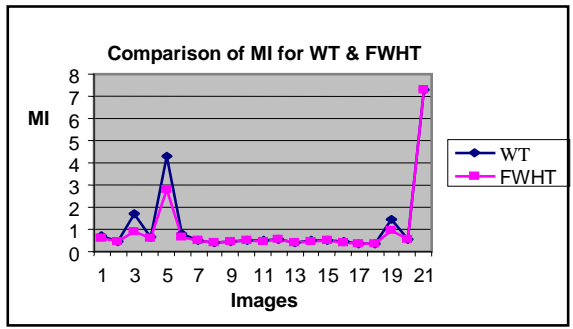

Figure 22: MRI T2-Registered –Sagittal Image 400 x 419  - 88.8kB
Using FWHT

Table 1.Represents results for WT, and FWHT using MI

| S.No | X in mm | Y in mm | Angle in degrees | MI after registration for WT | MI after registration for FWHT |
|---|---|---|---|---|---|
| 1 | 4 | -10 | 9 | 0.6760 | 0.5759 |
| 2 | -12 | -7 | 13 | 0.4560 | 0.4580 |
| 3 | 5 | -7 | 5 | 1.6951 | 0.8865 |
| 4 | -14 | -15 | 2 | 0.6655 | 0.5840 |
| 5 | -8 | -7 | 1 | 4.3194 | 2.7967 |
| 6 | 9 | 7 | -7 | 0.8229 | 0.6728 |
| 7 | 7 | -13 | 11 | 0.4955 | 0.4789 |
| 8 | 18 | 1 | 19 | 0.3766 | 0.3754 |
| 9 | -17 | 0 | -17 | 0.4577 | 0.4394 |
| 10 | 0 | -9 | 12 | 0.5064 | 0.4924 |
| 11 | 23 | -6 | 2 | 0.4982 | 0.4725 |
| 12 | -15 | 5 | -10 | 0.5726 | 0.5380 |
| 13 | 22 | 20 | 2 | 0.4061 | 0.4126 |
| 14 | 5 | 15 | 12 | 0.4790 | 0.4538 |
| 15 | -21 | 16 | -5 | 0.5023 | 0.5000 |
| 16 | -1 | 19 | 13 | 0.4330 | 0.4239 |
| 17 | 5 | 10 | -25 | 0.3673 | 0.3516 |
| 18 | -3 | 11 | 25 | 0.3426 | 0.3508 |
| 19 | 11 | -9 | 0 | 1.4506 | 0.9474 |
| 20 | 0 | 0 | 12 | 0.5513 | 0.5307 |
| 21 | 0 | 0 | 0 | 7.2952 | 7.2840 |

Figure 23.Comparison of WT and FWHT using MI.

Table 2.Represents results for WT, and FWHT using CC

| S.No | X in mm | Y in mm | Angle in degrees | CC after registration for WT | CC after registration for FWHT |
|---|---|---|---|---|---|
| 1 | 4 | -10 | 9 | 0.4840 | 0.4308 |
| 2 | -12 | -7 | 13 | 0.3170 | 0.3495 |
| 3 | 5 | -7 | 5 | 0.7801 | 0.6088 |
| 4 | -14 | -15 | 2 | 0.3876 | 0.4011 |
| 5 | -8 | -7 | 1 | 0.9425 | 0.9214 |
| 6 | 9 | 7 | -7 | 0.5282 | 0.4889 |
| 7 | 7 | -13 | 11 | 0.3227 | 0.3463 |
| 8 | 18 | 1 | 19 | 0.2199 | 0.2632 |
| 9 | -17 | 0 | -17 | 0.1774 | 0.1992 |
| 10 | 0 | -9 | 12 | 0.3317 | 0.3658 |
| 11 | 23 | -6 | 2 | 0.3952 | 0.4171 |
| 12 | -15 | 5 | -10 | 0.3131 | 0.3362 |
| 13 | 22 | 20 | 2 | 0.2667 | 0.3021 |
| 14 | 5 | 15 | 12 | 0.2765 | 0.3411 |
| 15 | -21 | 16 | -5 | 0.2638 | 0.3004 |
| 16 | -1 | 19 | 13 | 0.2377 | 0.3184 |
| 17 | 5 | 10 | -25 | 0.0987 | 0.1321 |
| 18 | -3 | 11 | 25 | 0.1537 | 0.2109 |
| 19 | 11 | -9 | 0 | 0.7487 | 0.6498 |
| 20 | 0 | 0 | 12 | 0.4324 | 0.4398 |
| 21 | 0 | 0 | 0 | 0.9965 | 0.9984 |

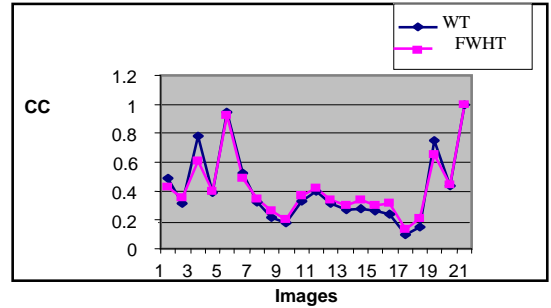

Figure 24.Comparison of WT and FWHT using CC.

Table 3.Represents Time consumption for Image Registration using WT and FWHT

| S.No | X in mm | Y in mm | Angle in degrees | Elapsed Time in seconds for WT | Elapsed Time in seconds for FWHT |
|---|---|---|---|---|---|
| 1 | 4 | -10 | 9 | 108.703000 | 3.812000 |
| 2 | -12 | -7 | 13 | 105.750000 | 3.734000 |
| 3 | 5 | -7 | 5 | 115.078000 | 3.844000 |
| 4 | -14 | -15 | 2 | 114.593000 | 3.844000 |
| 5 | -8 | -7 | 1 | 115.984000 | 3.937000 |
| 6 | 9 | 7 | -7 | 115.078000 | 3.750000 |
| 7 | 7 | -13 | 11 | 116.406000 | 3.766000 |
| 8 | 18 | 1 | 19 | 84.390000 | 3.797000 |
| 9 | -17 | 0 | -17 | 112.046000 | 3.718000 |
| 10 | 0 | -9 | 12 | 116.562000 | 3.781000 |
| 11 | 23 | -6 | 2 | 75.656000 | 3.719000 |
| 12 | -15 | 5 | -10 | 84.859000 | 3.813000 |
| 13 | 22 | 20 | 2 | 71.672000 | 3.750000 |
| 14 | 5 | 15 | 12 | 87.000000 | 3.781000 |
| 15 | -21 | 16 | -5 | 84.484000 | 3.797000 |
| 16 | -1 | 19 | 13 | 89.828000 | 3.735000 |
| 17 | 5 | 10 | -25 | 77.781000 | 3.703000 |
| 18 | -3 | 11 | 25 | 71.766000 | 3.687000 |
| 19 | 11 | -9 | 0 | 102.000000 | 3.907000 |
| 20 | 0 | 0 | 12 | 116.156000 | 3.766000 |
| 21 | 0 | 0 | 0 | 119.312000 | 3.766000 |

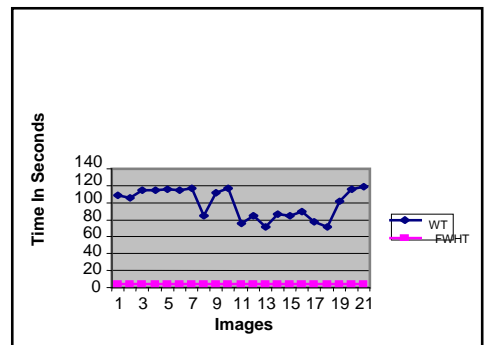

Figure 25.Comparison of WT and FWHT in terms of time

(b) For base 2:
For image 2

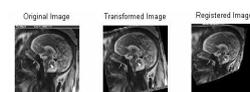

For image 7

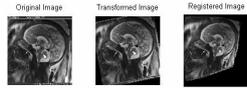

For image 8

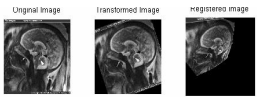

For image 9

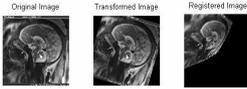

For image 10

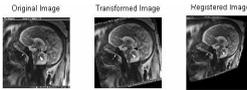

For image 11

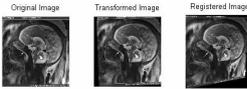

For image 12

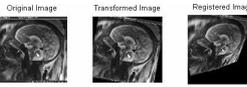

For image 13

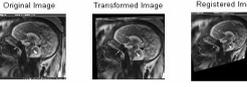

For image 14

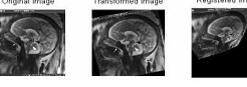

For image 15

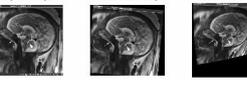

For image 16

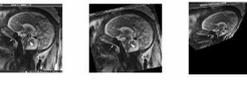

For image 17

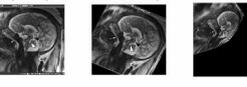

For image 18

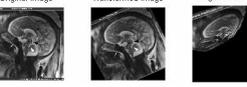

For image 20

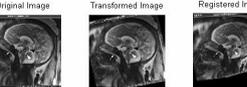

Figure 26: MRI T2-Registered –Sagittal Image 400 x 419 - 88.8kB using base 2 WT

Table 4.Represents results for base 2 of WT, and FWHT using MI

| S.No | X in mm | Y in mm | Angle in degrees | MI after registration for base 2 W T | MI after registration for base 2 FWHT |
|---|---|---|---|---|---|
| 1 | 4 | -10 | 9 | 4.3051 | 0.8614 |
| 2 | -12 | -7 | 13 | 0.4905 | 0.5424 |
| 3 | 5 | -7 | 5 | 4.3136 | 4.1158 |
| 4 | -14 | -15 | 2 | 4.3807 | 4.3630 |
| 5 | -8 | -7 | 1 | 4.3342 | 4.3213 |
| 6 | 9 | 7 | -7 | 4.3320 | 4.1178 |
| 7 | 7 | -13 | 11 | 0.7509 | 0.6762 |
| 8 | 18 | 1 | 19 | 0.5721 | 0.4058 |
| 9 | -17 | 0 | -17 | 0.5768 | 0.4637 |
| 10 | 0 | -9 | 12 | 0.6196 | 0.6446 |
| 11 | 23 | -6 | 2 | 0.9336 | 0.7394 |
| 12 | -15 | 5 | -10 | 0.5874 | 0.7129 |
| 13 | 22 | 20 | 2 | 0.7859 | 0.4227 |
| 14 | 5 | 15 | 12 | 0.6063 | 0.5107 |
| 15 | -21 | 16 | -5 | 0.6566 | 0.6576 |
| 16 | -1 | 19 | 13 | 0.5615 | 0.4770 |
| 17 | 5 | 10 | -25 | 0.5809 | 0.3608 |
| 18 | -3 | 11 | 25 | 0.5893 | 0.3776 |
| 19 | 11 | -9 | 0 | 7.3026 | 7.3031 |
| 20 | 0 | 0 | 12 | 0.6660 | 0.6725 |
| 21 | 0 | 0 | 0 | 7.2931 | 7.2887 |

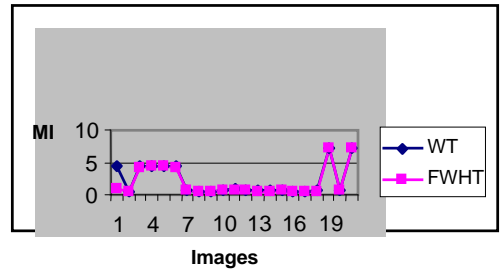

Figure 27.Comparison of base 2 of WT and FWHT using MI.

Table 5.Represents results for base 2 of WT, and FWHT using CC.

| S.No | X in mm | Y in mm | Angle in degrees | CC after registration for base 2 WT | CC after registration for base 2 FWHT |
|---|---|---|---|---|---|
| 1 | 4 | -10 | 9 | 0.8347 | 0.5752 |
| 2 | -12 | -7 | 13 | 0.0276 | 0.4258 |
| 3 | 5 | -7 | 5 | 0.8860 | 0.8864 |
| 4 | -14 | -15 | 2 | 0.8933 | 0.8934 |
| 5 | -8 | -7 | 1 | 0.9424 | 0.9426 |
| 6 | 9 | 7 | -7 | 0.8319 | 0.8329 |
| 7 | 7 | -13 | 11 | 0.1545 | 0.5183 |
| 8 | 18 | 1 | 19 | 0.0093 | 0.3054 |
| 9 | -17 | 0 | -17 | -0.0163 | 0.2186 |
| 10 | 0 | -9 | 12 | 0.1172 | 0.5127 |
| 11 | 23 | -6 | 2 | 0.3245 | 0.5808 |
| 12 | -15 | 5 | -10 | 0.0660 | 0.4894 |
| 13 | 22 | 20 | 2 | 0.1492 | 0.3133 |
| 14 | 5 | 15 | 12 | 0.0347 | 0.3929 |
| 15 | -21 | 16 | -5 | 0.1131 | 0.4107 |
| 16 | -1 | 19 | 13 | -0.0038 | 0.3385 |
| 17 | 5 | 10 | -25 | 0.0106 | 0.1531 |
| 18 | -3 | 11 | 25 | -0.0223 | 0.2464 |
| 19 | 11 | -9 | 0 | 0.9006 | 0.8985 |
| 20 | 0 | 0 | 12 | 0.1781 | 0.5338 |
| 21 | 0 | 0 | 0 | 0.9908 | 0.9981 |

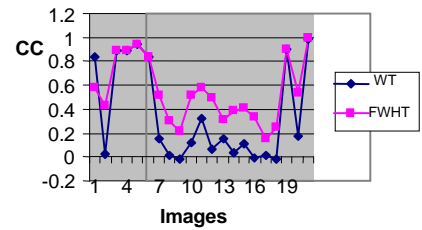

Figure 28.Comparison of base 2 of WT and FWHT using CC.

Table 6.Represents Time consumption for Image Registration using base 2 WT and FWHT

| S.No | X in mm | Y in mm | Angle in degrees | Elapsed Time in seconds for base 2 WT | Elapsed Time in seconds for base 2 FWHT |
|---|---|---|---|---|---|
| 1 | 4 | -10 | 9 | 141.734000 | 5.485000 |
| 2 | -12 | -7 | 13 | 64.141000 | 5.125000 |
| 3 | 5 | -7 | 5 | 144.906000 | 5.687000 |
| 4 | -14 | -15 | 2 | 143.781000 | 5.938000 |
| 5 | -8 | -7 | 1 | 144.125000 | 5.547000 |
| 6 | 9 | 7 | -7 | 138.172000 | 5.688000 |
| 7 | 7 | -13 | 11 | 109.234000 | 5.360000 |
| 8 | 18 | 1 | 19 | 67.360000 | 5.156000 |
| 9 | -17 | 0 | -17 | 54.594000 | 4.891000 |
| 10 | 0 | -9 | 12 | 106.625000 | 5.250000 |
| 11 | 23 | -6 | 2 | 120.781000 | 5.125000 |
| 12 | -15 | 5 | -10 | 86.922000 | 5.312000 |
| 13 | 22 | 20 | 2 | 89.688000 | 4.969000 |
| 14 | 5 | 15 | 12 | 75.234000 | 5.047000 |
| 15 | -21 | 16 | -5 | 82.547000 | 5.453000 |
| 16 | -1 | 19 | 13 | 55.360000 | 4.984000 |
| 17 | 5 | 10 | -25 | 55.672000 | 4.922000 |
| 18 | -3 | 11 | 25 | 65.484000 | 4.985000 |
| 19 | 11 | -9 | 0 | 134.703000 | 5.766000 |
| 20 | 0 | 0 | 12 | 110.781000 | 5.297000 |
| 21 | 0 | 0 | 0 | 144.875000 | 5.000000 |

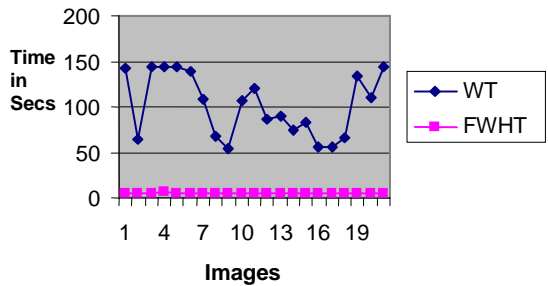

Figure 29.Comparison of base 2 of WT and FWHT in terms of time

(c) For base 5:

Table 7.Represents results for base 5 of WT, and FWHT using MI

| S.No | X in mm | Y in mm | Angle in degrees | MI after registration for base 5 WT | MI after registration for base 5 FWHT |
|---|---|---|---|---|---|
| 1 | 4 | -10 | 9 | 4.3363 | 4.0421 |
| 2 | -12 | -7 | 13 | 4.3264 | 4.0996 |
| 3 | 5 | -7 | 5 | 4.3136 | 4.1160 |
| 4 | -14 | -15 | 2 | 4.3802 | 4.3620 |
| 5 | -8 | -7 | 1 | 0.4197 | 4.3213 |
| 6 | 9 | 7 | -7 | 4.3315 | 4.1368 |
| 7 | 7 | -13 | 11 | 4.3147 | 4.2552 |
| 8 | 18 | 1 | 19 | 4.2660 | 3.6528 |
| 9 | -17 | 0 | -17 | 4.3429 | 4.3334 |
| 10 | 0 | -9 | 12 | 4.3727 | 4.0351 |
| 11 | 23 | -6 | 2 | 4.2871 | 4.1984 |
| 12 | -15 | 5 | -10 | 4.3386 | 4.2534 |
| 13 | 22 | 20 | 2 | 4.3571 | 4.0040 |
| 14 | 5 | 15 | 12 | 4.3256 | 4.0442 |
| 15 | -21 | 16 | -5 | 4.3478 | 3.9213 |
| 16 | -1 | 19 | 13 | 4.3553 | 3.9980 |
| 17 | 5 | 10 | -25 | 0.3670 | 2.7545 |
| 18 | -3 | 11 | 25 | error | 4.2565 |
| 19 | 11 | -9 | 0 | 7.3026 | 7.3028 |
| 20 | 0 | 0 | 12 | 4.3677 | 4.3685 |
| 21 | 0 | 0 | 0 | 7.2836 | 7.2923 |

For image 5

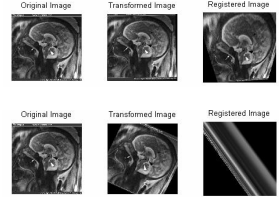

For image 17

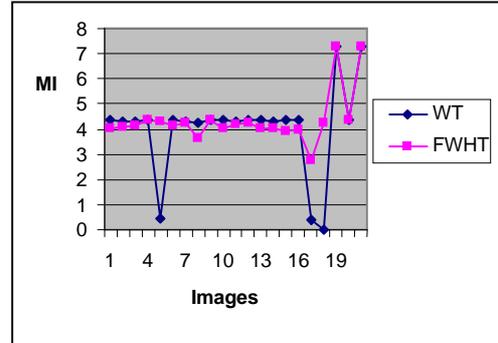

Figure 30: MRI T2-Registered –Sagittal Image 400 x 419 - 88.8kB using base 5 WT

Figure 31.Comparison of base 5 of WT and FWHT using MI.

Table 8.Represents results for base 5 of WT, and FWHT using CC.

| S.No | X in mm | Y in mm | Angle in degrees | CC after registration for base 5 WT | CC after registration for base 5 FWHT |
|---|---|---|---|---|---|
| 1 | 4 | -10 | 9 | 0.8342 | 0.8343 |
| 2 | -12 | -7 | 13 | 0.8459 | 0.8464 |
| 3 | 5 | -7 | 5 | 0.8860 | 0.8864 |
| 4 | -14 | -15 | 2 | 0.8934 | 0.8934 |
| 5 | -8 | -7 | 1 | 0.2565 | 0.9426 |
| 6 | 9 | 7 | -7 | 0.8319 | 0.8327 |
| 7 | 7 | -13 | 11 | 0.7713 | 0.7719 |
| 8 | 18 | 1 | 19 | 0.6627 | 0.6648 |
| 9 | -17 | 0 | -17 | 0.8052 | 0.8054 |
| 10 | 0 | -9 | 12 | 0.8206 | 0.8201 |
| 11 | 23 | -6 | 2 | 0.8006 | 0.8009 |
| 12 | -15 | 5 | -10 | 0.8408 | 0.8416 |
| 13 | 22 | 20 | 2 | 0.7378 | 0.7381 |
| 14 | 5 | 15 | 12 | 0.7152 | 0.7171 |
| 15 | -21 | 16 | -5 | 0.8282 | 0.8297 |
| 16 | -1 | 19 | 13 | 0.7201 | 0.7207 |
| 17 | 5 | 10 | -25 | 0.1778 | 0.6876 |
| 18 | -3 | 11 | 25 | error | 0.6574 |
| 19 | 11 | -9 | 0 | 0.9006 | 0.8969 |
| 20 | 0 | 0 | 12 | 0.8226 | 0.8226 |
| 21 | 0 | 0 | 0 | 0.9930 | 0.9911 |

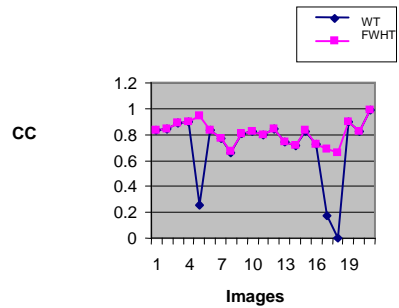

Figure 32.Comparison of base 5 of WT and FWHT using CC.

Table 9.Represents Time consumption for Image Registration using base 5 WT and FWHT

| S.No | X in mm | Y in mm | Angle in degrees | Elapsed Time in seconds for base 2 WT | Elapsed Time in seconds for base 2 FWHT |
|------|---------|---------|------------------|---------------------------------------|------------------------------------------|
| 1  | 4   | -10 | 9   | 147.125000 | 6.828000 |
| 2  | -12 | -7  | 13  | 136.250000 | 7.297000 |
| 3  | 5   | -7  | 5   | 141.453000 | 6.234000 |
| 4  | -14 | -15 | 2   | 136.797000 | 6.734000 |
| 5  | -8  | -7  | 1   | 138.813000 | 6.031000 |
| 6  | 9   | 7   | -7  | 139.047000 | 6.532000 |
| 7  | 7   | -13 | 11  | 135.640000 | 7.109000 |
| 8  | 18  | 1   | 19  | 131.563000 | 7.953000 |
| 9  | -17 | 0   | -17 | 132.546000 | 7.812000 |
| 10 | 0   | -9  | 12  | 134.750000 | 7.141000 |
| 11 | 23  | -6  | 2   | 137.797000 | 6.765000 |
| 12 | -15 | 5   | -10 | 137.234000 | 7.031000 |
| 13 | 22  | 20  | 2   | 134.969000 | 7.344000 |
| 14 | 5   | 15  | 12  | 134.609000 | 7.250000 |
| 15 | -21 | 16  | -5  | 134.985000 | 7.063000 |
| 16 | -1  | 19  | 13  | 135.687000 | 7.406000 |
| 17 | 5   | 10  | -25 | 76.594000  | 8.390000 |
| 18 | -3  | 11  | 25  | error      | 8.391000 |
| 19 | 11  | -9  | 0   | 151.719000 | 6.297000 |
| 20 | 0   | 0   | 0   | 133.453000 | 7.109000 |
| 21 | 0   | 0   | 0   | 150.047000 | 5.453000 |

For Image 5

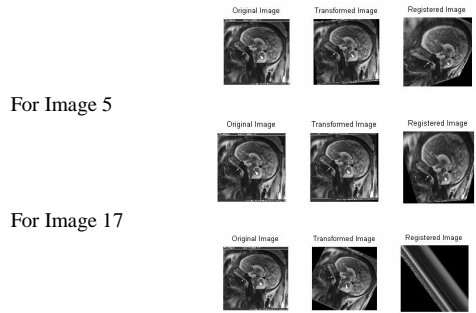

For Image 17

Figure 34: MRI T2-Registered –Sagittal Image 400 x 419 - 88.8kB using base 10 WT

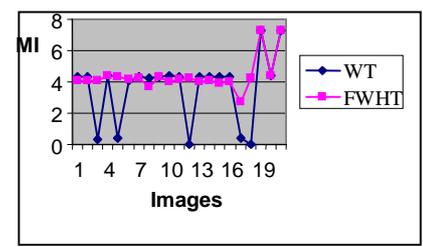

Figure 35.Comparison of base 10 of WT and FWHT using MI.

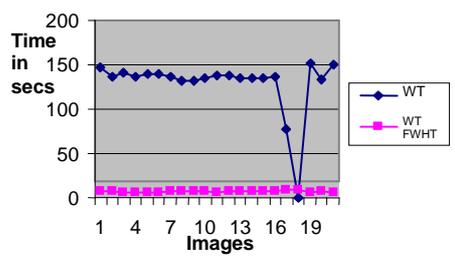

Figure 33.Comparison of base 5 of WT and FWHT in terms of time

**(d) For base 10:**

Table 10.Represents results for base 10 of WT, and FWHT using MI

| S.No | X in mm | Y in mm | Angle in degrees | MI after registration for base 5 WT | MI after registration for base 5 FWHT |
|------|---------|---------|------------------|--------------------------------------|----------------------------------------|
| 1  | 4   | -10 | 9   | 4.3362 | 4.0421 |
| 2  | -12 | -7  | 13  | 4.3256 | 4.0996 |
| 3  | 5   | -7  | 5   | 0.3304 | 4.1160 |
| 4  | -14 | -15 | 2   | 4.3722 | 4.3620 |
| 5  | -8  | -7  | 1   | 0.4205 | 4.3213 |
| 6  | 9   | 7   | -7  | 4.1077 | 4.1368 |
| 7  | 7   | -13 | 11  | 4.3106 | 4.2549 |
| 8  | 18  | 1   | 19  | 4.2627 | 3.6573 |
| 9  | -17 | 0   | -17 | 4.3432 | 4.3334 |
| 10 | 0   | -9  | 12  | 4.3724 | 4.0351 |
| 11 | 23  | -6  | 2   | 4.2861 | 4.1984 |
| 12 | -15 | 5   | -10 | error  | 4.2534 |
| 13 | 22  | 20  | 2   | 4.3571 | 4.0040 |
| 14 | 5   | 15  | 12  | 4.3263 | 4.0442 |
| 15 | -21 | 16  | -5  | 4.3480 | 3.9213 |
| 16 | -1  | 19  | 13  | 4.3539 | 3.9980 |
| 17 | 5   | 10  | -25 | 0.3677 | 2.7545 |
| 18 | -3  | 11  | 25  | error  | 4.2565 |
| 19 | 11  | -9  | 0   | 7.3026 | 7.3028 |
| 20 | 0   | 0   | 12  | 4.3679 | 4.3685 |
| 21 | 0   | 0   | 0   | 7.2892 | 7.2923 |

For Image 3

Table 11.Represents results for base 10 of WT, and FWHT using CC.

| S.No | X in mm | Y in mm | Angle in degrees | CC after registration for base 5 WT | CC after registration for base 5 FWHT |
|------|---------|---------|------------------|--------------------------------------|----------------------------------------|
| 1  | 4   | -10 | 9   | 0.8342 | 0.8343 |
| 2  | -12 | -7  | 13  | 0.8459 | 0.8464 |
| 3  | 5   | -7  | 5   | 0.0416 | 0.8864 |
| 4  | -14 | -15 | 2   | 0.8933 | 0.8934 |
| 5  | -8  | -7  | 1   | 0.2542 | 0.9426 |
| 6  | 9   | 7   | -7  | 0.8306 | 0.8327 |
| 7  | 7   | -13 | 11  | 0.7713 | 0.7719 |
| 8  | 18  | 1   | 19  | 0.6628 | 0.6645 |
| 9  | -17 | 0   | -17 | 0.8054 | 0.8054 |
| 10 | 0   | -9  | 12  | 0.8206 | 0.8201 |
| 11 | 23  | -6  | 2   | 0.8006 | 0.8009 |
| 12 | -15 | 5   | -10 | error  | 0.8416 |
| 13 | 22  | 20  | 2   | 0.7378 | 0.7381 |
| 14 | 5   | 15  | 12  | 0.7151 | 0.7171 |
| 15 | -21 | 16  | -5  | 0.8282 | 0.8297 |
| 16 | -1  | 19  | 13  | 0.7200 | 0.7207 |
| 17 | 5   | 10  | -25 | 0.1778 | 0.6876 |
| 18 | -3  | 11  | 25  | error  | 0.6574 |
| 19 | 11  | -9  | 0   | 0.9006 | 0.8969 |
| 20 | 0   | 0   | 12  | 0.8226 | 0.8226 |
| 21 | 0   | 0   | 0   | 0.9896 | 0.9911 |

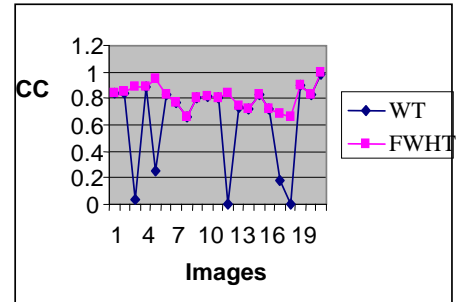

Figure 36.Comparison of base 10 of WT and FWHT using CC.

Table 12.Represents Time consumption for Image Registration using base 5 WT and FWHT

| S.No | X in mm | Y in mm | Angle in degrees | Elapsed Time in seconds for base 2 WT | Elapsed Time in seconds for base 2 FWHT |
|---|---|---|---|---|---|
| 1 | 4 | -10 | 9 | 138.297000 | 6.829000 |
| 2 | -12 | -7 | 13 | 135.922000 | 7.328000 |
| 3 | 5 | -7 | 5 | 133.406000 | 6.328000 |
| 4 | -14 | -15 | 2 | 136.000000 | 6.750000 |
| 5 | -8 | -7 | 1 | 141.125000 | 6.125000 |
| 6 | 9 | 7 | -7 | 139.000000 | 6.547000 |
| 7 | 7 | -13 | 11 | 136.703000 | 7.078000 |
| 8 | 18 | 1 | 19 | 131.750000 | 7.953000 |
| 9 | -17 | 0 | -17 | 132.828000 | 7.812000 |
| 10 | 0 | -9 | 12 | 135.328000 | 7.156000 |
| 11 | 23 | -6 | 2 | 138.907000 | 6.797000 |
| 12 | -15 | 5 | -10 | error | 6.968000 |
| 13 | 22 | 20 | 2 | 134.687000 | 7.344000 |
| 14 | 5 | 15 | 12 | 135.266000 | 7.219000 |
| 15 | -21 | 16 | -5 | 135.469000 | 7.109000 |
| 16 | -1 | 19 | 13 | 136.782000 | 7.375000 |
| 17 | 5 | 10 | -25 | 100.563000 | 8.343000 |
| 18 | -3 | 11 | 25 | error | 8.390000 |
| 19 | 11 | -9 | 0 | 142.500000 | 6.312000 |
| 20 | 0 | 0 | 12 | 134.797000 | 7.125000 |
| 21 | 0 | 0 | 0 | 149.781000 | 5.453000 |

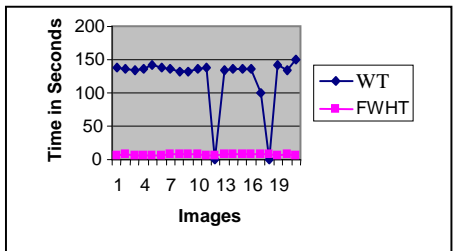

Figure 37.Comparison of base 10 of WT and FWHT in terms of time

From the above analysis it proves that the performance of the FWHT is better than the WT in terms of all the measures.

## IV. CONCLUSION

This paper proposes a new algorithm for medical image registration. A Fast Walsh Hadamard Transform is proposed in this paper for medical image registration. This transform reduces the time consumption in image registration. Therefore it proves to be a better approach for medical image registration than any other conventional Walsh Transform. The coefficients obtained using this transform are then normalized to obtain the unique number. This unique number represents the local structure of an image. Moreover this unique number indicates the feature of an image for image registration. The experimental results revealed the fact that the proposed algorithm using Fast Walsh Hadamard Transform performed well in image registration. The future work concentrates on further improvement in the results by using some other transforms that use correlation coefficients.

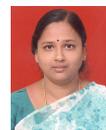
D.Sasikala is presently working as Assistant Professor, Department of CSE, Bannari Amman Institute of Technology, Sathyamangalam. She received B.E.( CSE) from Coimbatore Institute of Technology, Coimbatore and M.E. (CSE) from Manonmaniam Sundaranar University, Tirunelveli. She is now pursuing Phd in Image Processing. She has 11.5 years of teaching experience and has guided several UG and PG projects. She is a life member of ISTE. Her areas of interests are Image Processing, System Software, Artificial Intelligence, Compiler Design.

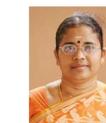
R. Neelaveni is presently working as a Assistant Professor, Department of EEE, PSG College of Technology, Coimbatore. She has a Bachelor's degree in ECE, a Master's degree in Applied Electronics and PhD in Biomedical Instrumentation. She has 23 years of teaching experience and has guided many UG and PG projects. Her research and teaching interests includes Applied Electronics, Analog VLSI, Computer Networks, and Biomedical Engineering. She is a Life member of Indian Society for Technical Education (ISTE). She has published several research papers in International, National Journals and Conferences.